\documentclass[11pt]{article}

\usepackage[utf8]{inputenc}
\usepackage{epsfig,amsmath,latexsym,amssymb}
\usepackage{graphicx}
\usepackage{lscape}
\usepackage{picture, eso-pic, tikz} 
\usepackage{dsfont}
\usepackage{listings}
\usepackage{changes}
\usepackage{hyperref}

\usepackage{algorithm,algorithmic}
\usepackage{subfigure}
\usepackage{natbib}
\usepackage{booktabs}
\usepackage{multirow}
\usepackage{tikz}
\usepackage{pgfplots}
\usetikzlibrary{positioning,arrows.meta,shapes.geometric}
\usepackage{float}
\def\R{{\mathbb R}}  
\def\N{{\mathbb N}}  
\def\Beweis{\footnotesize}

\newcommand{\Remm}[1]{}
\newtheorem{theo}{Theorem}[section]

\newtheorem{prop}[theo]{Proposition}

\newtheorem{model ass}[theo]{Model Assumptions}

\newtheorem{rem}[theo]{Remark}

\makeatletter
\def\thm@space@setup{%
  \thm@preskip=8pt plus 2pt minus 4pt
  \thm@postskip=\thm@preskip
}
\makeatother

\def\EndProof{\hfill {\scriptsize $\Box$}}

\numberwithin{equation}{section}

\definecolor{MyGray}{rgb}{0.92,0.92,0.92}


\newcommand{\red}[1]{\textcolor{red}{{#1}}}

\definecolor{BritishRacingGreen}{rgb}{0.0,0.5,0.0}
\def\bx{\boldsymbol{x}}

\def\bz{\boldsymbol{z}}

\def\b0{\boldsymbol{0}}

\def\bc{\boldsymbol{c}}
\def\bk{\boldsymbol{k}}
\def\bq{\boldsymbol{q}}
\def\bv{\boldsymbol{v}}

\def\bh{\boldsymbol{h}}

\def\bM{\mathbf{M}}
\def\bQ{\mathbf{Q}}
\def\bK{\mathbf{K}}
\def\bV{\mathbf{V}}
\def\bH{\mathbf{H}}
\def\bA{\mathbf{A}}
\def\bC{\mathbf{C}}

\title{In-Context Learning Enhanced Credibility Transformer}
\author{Kishan Padayachy\footnote{insureAI, kishan@insureai.co} \and
Ronald Richman\footnote{insureAI and University of the Witwatersrand, ronaldrichman@gmail.com}
\and Salvatore Scognamiglio\footnote{Department of Management and Quantitative Sciences, University of Naples ``Parthenope",\newline salvatore.scognamiglio@uniparthenope.it}
\and Mario V.~W\"uthrich\footnote{Department of Mathematics, ETH Zurich,
mario.wuethrich@math.ethz.ch}}

\date{\today}
\begin{document}

\maketitle

\begin{abstract}
The starting point of our network architecture is the Credibility Transformer which extends the classical Transformer architecture by a credibility mechanism to improve model learning and predictive performance. This Credibility Transformer learns credibilitized CLS tokens that serve as learned representations of the original input features. In this paper, we present a new paradigm that augments this architecture by an in-context learning mechanism, i.e., we increase the information set by a context batch of similar instances. This allows the model to enhance the CLS token representations of the instances by additional in-context information and fine-tuning. We empirically verify that this in-context learning improves predictive performance by adapting to similar risk patterns. Moreover, this in-context learning also allows the model to generalize to new instances which, e.g., have feature levels in the categorical covariates that have not been present when the model was trained -- for a relevant example, think of a new vehicle model which has just been launched by a car manufacturer.

\bigskip


\end{abstract}

\section{Introduction}
A fundamental challenge in actuarial modeling lies in balancing predictive performance, explainability and reliable model estimation. This is especially true for risks for which one does not have a long observation history.
Classical credibility theory, formalized by \citet{buhlmann1967experience} and \citet{buhlmann1970glaubwurdigkeit}, provides an elegant framework for addressing this challenge through an optimal linear combination of instance individual and collective claims experience. 
In parallel, the machine learning community has developed sophisticated approaches for learning from limited data, culminating in the emergence of in-context learning (ICL) capabilities in large-scale Transformer models. The discovery that large-scale Transformers can adapt to new tasks through demonstration examples, revolutionized our understanding of few-shot learning \citet{brown2020language}. That is, based on off-line training, a large-scale Transformer architecture can perform new tasks if sufficient context to these new tasks is provided. In particular, these tasks are done without retraining the Transformer architecture to the new situation. In some sense, this resembles the capability of being able to perform transfer learning. Recent theoretical work has established that this phenomenon can be understood as implicit gradient descent \citet{akyurek2023what} or Bayesian inference \citet{xie2022explanation}. For
ICL on tabular data, see \citet{hollmann2023} and \citet{muller2024}.

Recent work by \citet{richman2024credibility} introduced the Credibility Transformer, demonstrating how classical credibility concepts can be embedded within modern Transformer architectures helping to improve the predictive performance of Transformers on tabular input data. Their approach leverages the classify (CLS) token mechanism of \citet{devlin2019bert} to implement a learnable prior that is combined with feature-specific information through an attention-weighted averaging -- similarly to linear \citet{buhlmann1967experience} credibility. This Credibility Transformer architecture achieves state-of-the-art performance on insurance datasets while maintaining some explainability. A main benefit of the credibility mechanism (that acts similar to drop-out during model training) is the stabilization of the training algorithm leading to a delayed early stopping point and better feature extraction.

However, existing approaches, including the Credibility Transformer, evaluate each risk in isolation during inference, neglecting potentially valuable information from similar instances within the portfolio. This limitation becomes particularly relevant in practical scenarios where insurers must price policies for risk profiles with a limited historical experience while having access to recent claims data from similar policies, just think of expanding an existing insurance product to a new geographical region. The ability to leverage this contextual information dynamically, without retraining the model, may significantly enhance predictive accuracy and adapt to risk patterns of similar experience.

\subsection{Literature review and background}

\subsubsection{Classical credibility theory}
Credibility theory provides the theoretical foundation for combining individual experience with collective information in actuarial applications. The seminal work of \citet{buhlmann1967experience} and \citet{buhlmann1970glaubwurdigkeit}
established the mathematical framework for an optimal linear prediction taking into account both sources of information. In general, combining different sources of information involves intractable Bayesian posterior computations, and coping with this intractability, 
\citet{buhlmann1967experience} proposed a best-linear approximation by minimizing the mean squared error (MSE). 
This linear approximation takes the form
\begin{equation}\label{eq:classical_credibility}
\widehat{\mu} = \alpha Y + (1 - \alpha) \mu_0,
\end{equation}
where $Y$ reflects the individual experience, $\mu_0$ is the collective information, and $\alpha \in [0,1]$ is a {\it credibility weight} that weighs these two components. The credibility weight takes the form
\begin{equation}\label{BS credibility weight}
\alpha = \frac{v}{v + \kappa},
\end{equation}
where $v>0$ is a case weight indicating the amount of information contained in the individual observation $Y$ and $\kappa>0$ is the {\it credibility coefficient} (hyper-parameter) that balances between individual and collective information. This credibility coefficient $\kappa$ describes the trade-off between the different uncertainties involved at the individual and the collective risk level; see \citet{BuhlmannGisler} for an extended discussion.

\subsubsection{Transformer architectures for tabular data}

The adaptation of the Transformer architecture of \citet{vaswani2017attention}
 to tabular input data presents some challenges compared to their applications in large language models. Unlike sequential or grid-structured data, tabular features lack an inherent time-causal (positional) relationship for a canonical tensor encoding. This requires novel approaches to tokenization and positional encoding.
The TabTransformer  of \citet{huang2020tabtransformer} pioneered the application of Transformers to tabular data by applying self-attention exclusively to categorical embeddings while processing continuous features through a separate pathway. This design choice reflects the observation that categorical variables often exhibit complex interactions that benefit from attention mechanisms, while continuous features may adequately be handled by simpler feed-forward neural network (FNN) layers.

The feature tokenizer (FT) Transformer of  \citet{gorishniy2021revisiting} extends this approach by tokenizing all features, both categorical and continuous ones, and processing them through a unified Transformer architecture. Their continuous feature tokenization uses a simple FNN embedding with an embedding space dimension being identical to the one chosen for the categorical entity embeddings. This then naturally allows one to write the input feature in tensor form, which is the suitable structure for Transformer architectures. In an actuarial context, \citet{Brauer} has empirically verified the excellent performance of these FT Transformer architectures on tabular input data.
Recent work on numerical feature encoding has explored various alternative approaches to handle continuous variables in Transformers. Noteworthy, piecewise linear encoding (PLE) of \citet{gorishniy2022embeddings} provides a differentiable binning of continuous features that adapts to the data distribution. 

\subsubsection{In-context learning theory}

ICL emerged as a surprising capability of large language models, where models can adapt to new tasks through demonstration examples provided in the input, without any parameter updates and model retraining; see \citet{brown2020language}. This phenomenon has sparked intensive research into understanding the underlying mechanisms and theoretical foundations of ICL.
\citet{akyurek2023what} demonstrated that linear self-attention layers can implement gradient descent on a MSE objective function when provided with appropriate in-context examples. Their analysis shows that Transformers can internally implement optimization algorithms, with attention weights encoding gradient information and value projections performing parameter updates. This finding suggests that ICL implements a form of implicit meta-learning within the forward pass.
From a Bayesian perspective, \citet{xie2022explanation} and \citet{muller2024}
interpret ICL as implicit Bayesian inference, where the model maintains a posterior distribution over possible tasks and updates this posterior based on the context examples. Under this framework, the pre-trained distribution defines the prior, and attention mechanisms implement approximate posterior inference, providing a probabilistic foundation for understanding ICL behavior. 

Recent work has identified important factors that influence ICL performance.  \citet{min2022rethinking} demonstrate that the format and distribution of demonstration examples matter more than their correctness, suggesting that ICL relies heavily on pattern matching and format recognition, while  \citet{wei2023larger} show that ICL capabilities emerge at scale, with larger models exhibiting qualitatively different learning behaviors than smaller ones.

Within the tabular learning context, where models are usually trained until performance on a validation loss is optimal, it would be surprising to find that ICL could lead to additional significant improvements in performance and -- if this were to be the case -- a new explanation for this improvement would be needed. Whereas \citet{muller2024} have indeed shown that their TabICL model outperforms many other models, including models that only use supervised learning, as well as those using ICL, nonetheless, this could be explained by their extensive foundation model pre-training scheme using real and synthetic data. 
Classical supervised learning models are typically trained on a finite sample using early stopping of the gradient descent algorithm to prevent over-fitting. This early stopping may also prevent learning a more detailed, fine-grained structure, and foundation models being trained on a large body of (similar) data may be aware of such a more detailed structure.

\subsubsection{Connections between credibility and meta-learning}

The connection between actuarial credibility theory and modern meta-learning approaches has received limited attention despite their conceptual similarities. Both frameworks address the fundamental challenge of making predictions with limited task-specific data by leveraging broader experience. In credibility theory, this manifests as combining individual risk experience with portfolio-level information. In meta-learning, it involves leveraging experience across multiple tasks to enable rapid adaptation to new tasks.
The mathematical structures underlying these approaches exhibit close parallels. The linear credibility formula \eqref{eq:classical_credibility} can be seen as a special case of prototype-based meta-learning, where the collective mean $\mu_0$ serves as a prototype and the credibility factor $\alpha$ determines the adaptation strength. Similarly, the attention mechanisms in Transformers implement a form of non-parametric regression that generalizes attention weighting to high-dimensional feature spaces.
Recent work in neural processes \citep{garnelo2018neural} provides a framework that unifies these perspectives. Neural processes implement a form of meta-learning that closely resembles Bayesian inference, with the context set defining a posterior distribution over functions. This framework naturally accommodates varying amounts of context information, analogous to how credibility factors adjust (based on available experience) through the case weights $v>0$, see \eqref{BS credibility weight}.

\subsection{In-context learning enhanced Credibility Transformer}
Our main contribution is the introduction of the In-Context Learning enhanced Credibility Transformer (ICL-Credibility Transformer), which
combines the methods discussed in the previous subsections. First, the Credibility Transformer of \citet{richman2024credibility} leverages the FT Transformer with a credibility mechanism that significantly enhances model training. A crucial ingredient of the Credibility Transformer is the inclusion of the CLS token of \citet{devlin2019bert}, that is used to learn an encoding of the input information using the classical attention layer mechanism of \citet{vaswani2017attention}. Instead of simply encoding and extracting this CLS token, it is credibilitized with a linear credibility mechanism that combines the feature individual CLS token encoding with a collective information token; see \citet{richman2024credibility}. This linear credibility combination has a very positive effect on gradient descent training, resulting in excellent predictive models. In fact, during gradient descent training, a randomized version of the credibility mechanism is applied that either selects the instance-specific CLS token or the collective information token. This randomized mechanism during model training can be interpreted as a drop-out mechanism, similar to \citet{srivastava2014dropout} and \citet{wager2013dropout}, but instead of just dropping out by masking the information, the Credibility Transformer rather learns a global representation (similar to the collective mean $\mu_0$, see \eqref{eq:classical_credibility}) on the dropped out instances. This construction results in the credibilitized CLS token which is a robustified representation of the input data. The Credibility Transformer then processes these tokens through a decoder forming the predictions.

Inspired by the ICL architectures of \citet{hollmann2023} and \citet{muller2024} on tabular data, we modify the Credibility Transformer architecture to facilitate ICL. 
This raises the following architectural questions:
\begin{itemize}
\item At which stage should the predictive model on individual instances be augmented by context information? 
\item How can this augmentation technically be done?
\item How can we ensure robustness of model learning?
\end{itemize}

Our primary contributions are the answers to the above questions.
First, we provide an architectural innovation that enables ICL within the Credibility Transformer architecture. Second, we verify that this ICL improves the performance of the Credibility Transformer, and we give explanations for these improvements. Third, we show how the ICL-Credibility Transformer enables us to predict on new data with new categorical levels that have not been available during model training. The following items are going to be presented:

\paragraph{Architectural innovation.} We introduce the ICL-Credibility Transformer, a novel architecture that extends the {\it base Credibility Transformer} with ICL capabilities. Our design features the following three ICL components that serve at augmenting the credibilitized CLS tokens by context information including the response information of the context batch. 
\begin{itemize}
\item[(a)] We design an {\it outcome token decorator} which augments the CLS tokens of the {\it context batch} by response information
\item[(b)] We design a cross-batch attention mechanism that enables the instances of the {\it target batch} (to be predicted) to learn from the outcome augmented instances of the context batch. This context learning takes place in the space of the CLS tokens, i.e., in the same space as the original CLS tokens of the base Credibility Transformer live. This is enforced by reusing the frozen decoder from the base Credibility Transformer, as we discuss next.
\item[(c)] We use the same projection layer as in the base Credibility Transformer to decode the outcome decorated CLS tokens. This is justified by the fact mentioned in item (b) saying that the ICL takes place in the space of the CLS tokens, we come back to this in the training methodology below
\end{itemize}

\paragraph{Theoretical framework.} We establish formal connections between ICL mechanisms and classical credibility theory, proving that our ICL represents a credibility mechanism that regulates the impact of the context on the target instances to be predicted. We demonstrate that the attention weights serve as adaptive, data-driven credibility factors that depend on feature similarity, generalizing the fixed credibility weights of classical approaches.

\paragraph{Training methodology.} We deploy a two-phase training strategy that first establishes strong feature representations through a standard supervised learning method of the base Credibility Transformer. It has been verified in 
\citet{richman2024credibility} that this learned base Credibility Transformer already presents a strong predictive model. The second training phase fine-tunes the model using ICL. During this phase, we freeze the decoder layers (i.e., the output decoder) and train the ICL mechanism, as well as the base Credibility Transformer encoder. This constrained training approach enhances the robustness of the learning process, while allowing the weights of the base Credibility Transformer encoder (which were initially optimized for supervised learning) to be modified. Keeping the output decoder frozen acts as an information bottleneck ensuring that the ICL enhanced CLS tokens remain in the same encoding space guaranteeing robustness in this second fitting step.
Optionally, a third training step can be applied by jointly fine-tuning all network components, i.e., including the decoder layer; however, this requires careful regularization and a small learning rate to prevent performance degradation.

\paragraph{Empirical validation.} We verify on the popular French Motor Third Party Liability (MTPL) dataset of \citet{Dutang} the outstanding performance of the ICL-Credibility Transformer. Our study quantifies the contribution of each architectural component, while explainability analyses reveal that the learned attention patterns align with actuarial intuition about risk similarity and credibility weighting. An interesting property of ICL is that we can explicitly illustrate and try to understand how individual instances in the context batch enhance the predictions; we provide empirical results explaining this case by case. A second interesting case study shows how the ICL-Credibility Transformer can be used to predict on data with new covariate levels, e.g., a new region that has not been available during model training. This generalization property (similar to transfer learning) especially draws on the context, to find the right level of integration into the existing model.

\subsection{Manuscript organization}
The next section introduces the ICL-Credibility Transformer. Starting from the base Credibility Transformer, we augment this base architecture by ICL that modifies the instance representations in the latent CLS token space by context batch information. This is achieved by a context token decorator and an ICL transformer. We also discuss the two phase learning procedure, which is common practice in contemporary machine learning tools. Section \ref{section interpretation} mainly serves at giving model interpretations and making the connection to B\"uhlmann credibility. In Section \ref{Numerical example}, we verify the excellent performance of the ICL-Credibility Transformer on the popular French MTPL data of \cite{Dutang}. In Section \ref{Zero-shot analysis of Region}, we discuss a zero-shot analysis that shows how our method generalizes to new levels of features that have not been available for training. Finally, Section \ref{Conclusions} concludes.

\section{In-context learning enhanced Credibility Transformer}
This section defines the architecture of the ICL-Credibility Transformer. 
This architecture is crucially based on the groundbreaking work of
\cite{vaswani2017attention} on the attention layer mechanism and the Transformer architecture. Based on this seminal work, 
\cite{richman2024credibility} introduced the Credibility Transformer which adapts the classical Transformer to tabular data and equips the architecture with a credibility mechanism that enhances learning and prediction. This outlline is based on the notation of \cite{richman2024credibility}.

\subsection{ICL-Credibility Transformer architecture}

The ICL-Credibility Transformer architecture consists of four main components: 
\begin{itemize}
\item[(1)] a {\it Credibility Transformer encoder with a CLS token} that processes tabular features;
\item[(2)] an {\it outcome token decorator} that encodes claims information of the context batch; 
\item[(3)] an {\it ICL Transformer layer} that performs causal cross-batch attention; and 
\item[(4)] a {\it (frozen) decoder} that produces the final predictions from the tokens. 
\end{itemize}

We describe each component in detail in the following subsections.

\subsubsection{Credibility Transformer encoder with CLS token}
Our base Credibility Transformer (CT) follows the architecture introduced by  \cite{richman2024credibility}, and we also use the notation of that reference. It takes an input $\bx$ and it outputs a prediction $\mu^{\rm CT}(\bx)$ by an encoder-decoder architecture. The encoder $\bc^{\rm cred}$ takes an  input
$\bx$ and maps it to the credibilitized CLS token
\begin{equation*}
\bx ~\mapsto ~ \bc^{\rm cred}=\bc^{\rm cred}(\bx) \in \R^{2b},
\end{equation*} 
for a fixed encoding dimension $2b \in \N$;
see \citet[Formula (2.12)]{richman2024credibility}. This credibilitized CLS token is a $2b$-dimensional encoding of the feature $\bx$ (containing positional encodings). It is obtained by a Transformer architecture which is extended by the CLS token of Devlin et al.~\cite{devlin2019bert}. During model fitting this CLS token is computed by a credibility weighted average of individual instance information and global (prior) population information; for full technical details we refer to \citet[Section 2.2]{richman2024credibility}.

The second part of the Credibility Transformer is the decoder $\bz^{\rm decod}$
which maps the credibilitized CLS token to the prediction, usually we take a FNN architecture for the decoding step. Composing this encoder-decoder pair provides the base Credibility Transformer
\begin{equation}\label{base credibility transformer}
\bx ~\mapsto ~ \mu^{\rm CT}(\bx)
= \left(\bz^{\rm decod} \circ \bc^{\rm cred}\right)(\bx).
\end{equation}
This base Credibility Transformer architecture is described in full detail in \citet[Section 2]{richman2024credibility}.

\medskip

{\bf Training phase 1.}
This base Credibility Transformer \eqref{base credibility transformer} presents a supervised learning model. It is trained in a classical supervised learning manner by minimizing a loss function that compares the predictions
$\mu^{\rm CT}(\bx_i)$ to the outcomes $Y_i$ on an available learning sample 
${\cal L}=(Y_i,\bx_i,v_i)_{i=1}^n$; the variables $v_i>0$ present case weights that may differ across the instances. 
This supervised learning step provides us with a fitted encoder $\widehat{\bc}^{\rm cred}$, a fitted decoder $\widehat{\bz}^{\rm decod}$ and a predictive model
\begin{equation}\label{base credibility transformer estimate}
\bx ~\mapsto ~ \widehat{\mu}^{\rm CT}(\bx)
= \left(\widehat{\bz}^{\rm decod} \circ \widehat{\bc}^{\rm cred}\right)(\bx).
\end{equation}
This is precisely the predictive model considered
in \citet[Section 2]{richman2024credibility}, and in Table 2 of that reference  verifies the excellent predictive performance of this architecture on the popular French MTPL insurance dataset.

\medskip

Broadly speaking, for the following steps, we keep the decoder architecture $\widehat{\bz}^{\rm decod}$  of the trained base Credibility Transformer \eqref{base credibility transformer estimate} fixed, i.e., we freeze the estimated weights of the fitted decoder $\widehat{\bz}^{\rm decod}$, and before applying this frozen decoder $\widehat{\bz}^{\rm decod}$ on the credibilitized CLS tokens $\widehat{\bc}^{\rm cred}=\widehat{\bc}^{\rm cred}(\bx)$, we enrich these by in-context information. 

\subsubsection{Outcome token decorator}
In the next step, we insert a module into the pre-calibrated base Credibility Transformer
\eqref{base credibility transformer estimate} that enables ICL.
For this additional step, we assume that there two kinds of data available, namely, we assume having a {\it context batch} and a {\it target batch}. The target batch collects the instances whose responses need to be predicted. This prediction is supported not only by the features of the target batch, but also by context data for which the responses are available.

We denote the context batch by ${\cal B}_{\rm context}=(Y_j,\bx_j,v_j)_{j \in {\cal I}_{\rm context}}$, and the target batch is denoted by ${\cal B}_{\rm target}=(Y_i,\bx_i,v_i)_{i \in {\cal I}_{\rm target}}$. These two sets are {\it disjoint} to avoid any leakage of information. The outcome token decorator is used to enrich the credibilitized CLS tokens $\widehat{\bc}^{\rm cred}(\bx_j)$ of the context set
$j\in {\cal I}_{\rm context}$ by outcome (response) information $Y_j$, which is assumed to be available only on the context batch. This will then serve as the context to predict the responses in the target batch ${\cal B}_{\rm target}$. 

To simplify the notation, we merge the two index sets ${\cal I}=
{\cal I}_{\rm target} \cup {\cal I}_{\rm context}$ assuming that these indices allow for a unique identification of all instances. We then introduce the mask
\begin{equation*}
M_{i} = \begin{cases}
0 & \text{if $i \in {\cal I}_{\rm target}$,} \\
1 & \text{if $i \in {\cal I}_{\rm context}$.}
\end{cases}
\end{equation*}
This allows us to rewrite target and context batches in a unified batch
\begin{equation*}
{\cal B}=(Y_i,\bx_i,v_i, M_i)_{i \in {\cal I}},
\end{equation*}
with response $Y_i$ only being available for $M_i=1$.

\begin{figure}[htb!]
    \centering
    \includegraphics[width=.8\linewidth]{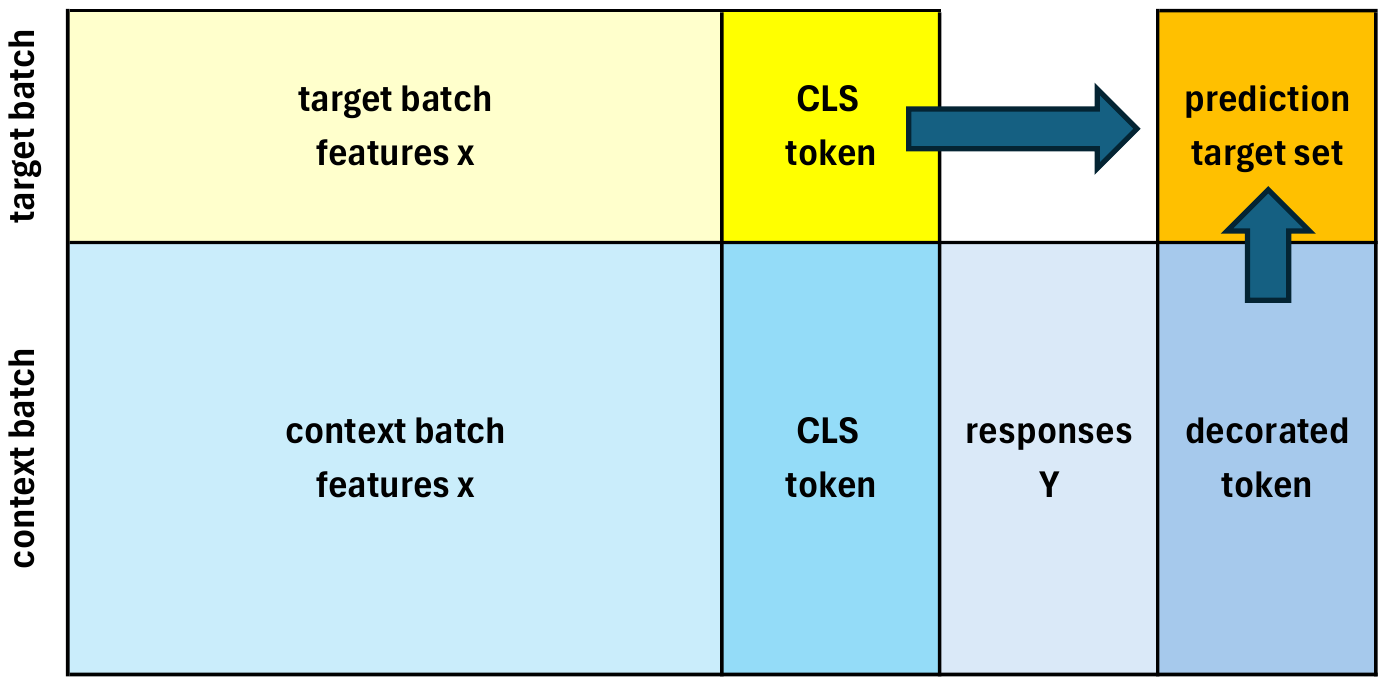}
    \vspace{-2.8cm}
    \caption{ICL-Credibility Transformer architecture processing the target batch ${\cal B}_{\rm target}$ (yellow) and the context batch ${\cal B}_{\rm context}$ (blue) to form the prediction on the target set (orange). }
    \label{fig: ICL-CT}
\end{figure}

The following {\it outcome token decorator} only acts on the context batch, i.e., on the instances $i\in {\cal I}$ with $M_i=1$. Let $\bz^{\rm FNN1}:\R \to \R^{2b}$ be a FNN network that embeds the responses into the same space as the credibilitized tokens
\begin{equation}\label{def: output token decorator}
Y ~\mapsto ~\bz^{\rm FNN1}(Y) \in \R^{2b}.
\end{equation}
This outcome decorator will enrich the representations with claim information for instances in the context batch while maintaining causal consistency. The weights of this outcome token decorator $\bz^{\rm FNN1}$ are learned in the second training phase, and this token decorator considers the unscaled responses $Y$ not involving the case weights $v$, to avoid any leakage of information through the case weights. That is, in case of claims counts $N$,
we simply select $Y=N$, which is achieved by setting $v \equiv 1$. 

\medskip

The {\it decorated token} is then defined by
\begin{equation}\label{decorated tokens}
\bc^{\rm decor}(\bx_i)
=\widehat{\bc}^{\rm cred}(\bx_i) + \frac{v_i}{v_i + \kappa} \, M_i \,\bz^{\rm FNN1}(Y_i),
\end{equation}
with
\begin{itemize}
\item credibilitized CLS token
$\widehat{\bc}^{\rm cred}(\bx_i)$, which, as above, has been produced using the encoder part of the base Credibility Transformer;
\item we decorate all instances $i \in {\cal I}_{\rm context}$ from the context batch ${\cal B}_{\rm context}$, $M_i=1$, with encoded target information $\bz^{\rm FNN1}(Y_i)$;
\item we integrate this target information using \cite{buhlmann1967experience} credibility, assigning a higher credibility for bigger case weights $v_i>0$, and the credibility coefficient $\kappa>0$ regulates the level of information considered; $\kappa$ is a hyper-parameter that is either chosen by the modeler or optimized by grid search.
\end{itemize}

\begin{rem}\normalfont
The output token decorator \eqref{def: output token decorator} takes as inputs the unscaled responses $Y$. This will be suitable for our application which will consider low frequency claim counts, but it may need to be adapted in high frequency or high weight situations to weight-scaled versions. The credibility weights are adjusted such that we give lower credibility to observations with lower weights, emphasizing that these observations contain more uncertainty (irreducible risk from pure randomness). Moreover, for other applications, e.g., claim size modeling, one may also use the residuals $Y - \widehat{\mu}^{\rm CT}(\bx)$ as inputs to the output token decorator. A pre-processing stage of the outcome token decorator should also consider transforming heavy-tailed or skewed responses, e.g., applying a log-transformation to heavy-tailed claim amounts, for a more efficient model training. 
\end{rem}

\subsubsection{ICL Transformer}
We are now equipped with the decorated tokens for all instances $i \in {\cal B}$ in the target and context batches, given by
\begin{equation*}
\bc^{\rm decor}(\bx_i) = \begin{cases}
\widehat{\bc}^{\rm cred}(\bx_i)  & \text{if $i \in {\cal I}_{\rm target}$,} \\
\widehat{\bc}^{\rm cred}(\bx_i) + \frac{v_i}{v_i + \kappa} \,\bz^{\rm FNN1}(Y_i) & \text{if $i \in {\cal I}_{\rm context}$,}
\end{cases}
\end{equation*}
we also refer to Figure \ref{fig: ICL-CT}.
The goal is to predict all responses $Y_i$ of the target set $i\in {\cal I}_{\rm target}$ in one shot using the entire context batch ${\cal B}_{\rm context}$. To ensure independence within the target set in the next step, we introduce the mask operator (on the log-scale)
\begin{equation*}
M^{\infty}_{i,j} =
\begin{cases}
-\infty & \text{if $i,j \in {\cal I}_{\rm target}$ and $i \neq j$},\\
0 & \text{otherwise.} \\
\end{cases}
\end{equation*}
This masks all pairs $(i,j)$ in the target batch ${\cal B}_{\rm target}$ with $-\infty$ if their indices $i\neq j$ differ. This will imply that the instances in the target batch cannot interact. We define the mask tensor $\bM =(M^\infty_{i,j})_{i,j \in {\cal I}}$, and we set
$n=|{\cal I}|$ for the total batch size and $n_1=|{\cal I}_{\rm target}|$ for the target batch size.

Next, we define the causal ICL Transformer layer.
We start by defining a causal attention layer; see \cite{vaswani2017attention}. Using three different time-distributed FNNs, we generate the query, key and value tensors, respectively,
\begin{eqnarray*}
\bQ &=& \left[\boldsymbol{z}^{\rm FNN}_{Q}\left(\bc^{\rm decor}(\bx_i)\right)\right]_{i\in {\cal I}}^\top~\in~\R^{n \times 2b},\\
\bK &=& \left[\boldsymbol{z}^{\rm FNN}_{K}\left(\bc^{\rm decor}(\bx_i)\right)\right]_{i\in {\cal I}}^\top~\in~\R^{n \times 2b},\\
\bV &=& \left[\boldsymbol{z}^{\rm FNN}_{V}\left(\bc^{\rm decor}(\bx_i)\right)\right]_{i\in {\cal I}}^\top~\in~\R^{n \times 2b},
\end{eqnarray*}
for three time-distributed FNNs $\boldsymbol{z}^{\rm FNN}_{\ell}$ having an identical architecture but different network weights for $\ell \in \{Q,K,V\}$. These three tensors are used to define the {\it causal self-attention head}
\begin{equation}\label{causal self-attention head}
\bH = \bH(\bQ,\bK,\bV,\bM) = \bA\, \bV ~\in ~\R^{n \times 2b},
\end{equation}
with {\it causal attention layer}
\begin{equation}\label{time-causal learning}
\bA=\bA(\bQ,\bK,\bM) = \operatorname{softmax} \left( \bQ \bK^\top/\sqrt{2b} + \bM \right)
~\in ~\R^{n \times n},
\end{equation}
where the softmax operator is applied row-wise.
This allows us to define the {\it ICL Transformer layer} as follows. First, we define the decorated token tensor
\begin{equation*}
\bC^{\rm decor}= \left[\bc^{\rm decor}(\bx_i)\right]_{i\in {\cal I}}^\top~\in~\R^{n \times 2b}.
\end{equation*}
The ICL Transformer layer includes two skip connections, another time-distributed FNN 
$\boldsymbol{z}^{\rm t\text{-}FNN2}$, and dropout layers and layer-normalization layers. The first regularized skip connection is given by
\begin{equation*}
\bC^{\rm mid} =  \operatorname{LayerNorm}\left(\bC^{\rm decor} +
 \operatorname{Dropout}(\bH)\right).
 \end{equation*}
 This then allows one to compute the ICL Transformer tensor
\begin{equation}\label{ICL Transformer output}
\bC^{\rm ICL\text{-}trans} =  \operatorname{LayerNorm}\left(\bC^{\rm mid} +
 \operatorname{Dropout}\left(\boldsymbol{z}^{\rm t\text{-}FNN2}\left(\bC^{\rm mid}\right)\right)\right) ~\in ~\R^{n \times 2b}.
\end{equation}
Remark that \eqref{ICL Transformer output} is the classical Transformer encoder introduced by \cite{vaswani2017attention}, and the mask operator $\bM$ in
\eqref{time-causal learning} and the time-distributed FNN layers ensure that different instances of the target batch do not interact in this ICL Transformer encoder.

\subsubsection{Frozen decoder}
The last step is to decode the ICL Transformer tensor $\bC^{\rm ICL\text{-}trans}$
to form the predictions on the target batch, i.e., for the responses $Y_i$, $i\in {\cal I}_{\rm target}$. For this we use the frozen decoder $\widehat{\bz}^{\rm decod}$
 from the first training phase.
 This architectural constraint ensures that all ICL computations operate within the original representation space learned during the first learning phase. This provides the practical benefit of maintaining the interpretability of the original CLS token space.

Let us denote the row vectors of the ICL Transformer tensor $\bC^{\rm ICL\text{-}trans}$ as follows
\begin{equation*}
\bC^{\rm ICL\text{-}trans}= \left[\bc^{\rm ICL\text{-}trans}_1,
\ldots, \bc^{\rm ICL\text{-}trans}_{n_1},\,\bc^{\rm ICL\text{-}trans}_{n_1+1},
\ldots, \bc^{\rm ICL\text{-}trans}_{n}\right]^\top  ~\in ~\R^{n \times 2b},
\end{equation*}
where the first $n_1=|{\cal I}_{\rm target}|$ rows belong to the target batch ${\cal B}_{\rm target}$ whose responses $Y_i$, $i\in {\cal I_{\rm target}}$, we would like to predict, and the remaining rows correspond to the context batch.
Applying the frozen decoder provides us in analogy to \eqref{base credibility transformer estimate} the decoder
\begin{equation}\label{ICL transformer}
\widehat{\mu}^{\rm ICL\text{-}CT}(\bx_i; {\cal B}_{\rm context})
= \widehat{\bz}^{\rm decod} \left( \bc^{\rm ICL\text{-}trans}_{i}  \right),
\qquad \text{ for $i \in {\cal I}_{\rm target}$}.
\end{equation}
We give remarks that all relate to the frozen decoder.
\begin{itemize}
\item {\it Calibration preservation:} The output scale and the calibration learned during the first training phase are preserved by keeping the weights of the decoder fixed during the second training phase (being discussed below).
\item {\it Implicit regularization:} By constraining the ICL mechanism to operate within the learned representation space, we prevent it from arbitrarily transforming representations in ways that might over-fit to the ICL training objective, i.e., this acts as a regularization mechanism. This can also be seen as an information bottleneck that prevents the ICL mechanism from encoding arbitrary context information.
\item {\it Transfer learning efficiency:} The frozen decoder ensures that the ICL mechanism leverages the full predictive power of the pre-trained model of the first training phase rather than relearning the entire prediction function. The ICL components only need to learn the context-aware adjustments within the existing representation space.
\item {\it Gradient flow simplification:} During the ICL fine-tuning (second training phase), the gradients flow through the ICL Transformer and the outcome token decorator, but  they stop at the frozen decoder, simplifying the optimization landscape and reducing the risk of forgetting of the pretrained representations.
\end{itemize}

\medskip

{\bf Training phase 2.} This second training phase is essentially an ICL fine-tuning. This ICL fine-tuning acts on
\begin{enumerate}
\item the encoder of the base Credibility Transformer;
\item the outcome token decorator \eqref{decorated tokens}, this involves the response embedding $\bz^{\rm FNN1}$ and the credibility coefficient $\kappa$;
\item the ICL Transformer, this involves the Transformer architecture \eqref{ICL Transformer output} including the attention mechanism and the skip connections architecture.
\end{enumerate}

The outputs of the ICL-Credibility Transformer architecture are the predictions
\eqref{ICL transformer} on the target batch. This motivates to consider
the loss function for the second training phase
\begin{equation}\label{ICL losses}
{\cal L}_{\text{ICL}}\left({\cal B}_{\rm target};{\cal B}_{\rm context}\right) 
 = \frac{1}{|{\cal I}_{\rm target}|} \sum_{i \in {\cal I}_{\rm target}} v_i \, L\left(Y_i, 
 \widehat{\mu}^{\rm ICL\text{-}CT}(\bx_i; {\cal B}_{\rm context})\right),
 \end{equation}
for a suitable loss function $L$ (comparing the responses $Y$ and predictions $\widehat{\mu}$) and case weights $v>0$.
In practice, we will have many target-context pairs, and this second training phase performs gradient descent fine-tuning on all these pairs, averaging
the losses \eqref{ICL losses} of the available target-context pairs.

\subsection{Learning procedure}
The fitting procedure presented above has two different learning phases. Phase 1: The first phase is used to learn the credibilitized CLS token encoder $\widehat{\bc}^{\rm cred}$ and the decoder $\widehat{\bz}^{\rm decod}$, respectively. Phase 2: The second phase is used to learn the context augmentation involving the context token decorator and the ICL Transformer, as well as fine-tuning the CLS token decorator. This two step approach is employed for robustifying the training procedure, but it also has computational advantages.
There is one critical point in this two stage procedure though, namely, one needs to ensure that the learned credibilitized CLS tokens $\widehat{\bc}^{\rm cred}(\bx_i)$, $i \in {\cal I}$, 
are sufficiently rich and carry sufficient information of the inputs $\bx_i$; an encoding can be thought of a data compression. The critical point is that the training of the encoder $\widehat{\bc}^{\rm cred}$ and the decoder 
$\widehat{\bz}^{\rm decod}$ only focuses on optimal predictive performance, without accounting for the space of encoded features $(\widehat{\bc}^{\rm cred}(\bx))_{i\in {\cal I}}$ being still sufficiently rich to be able to benefit from the ICL mechanism in the second step. Basically, there are two ways of ensuring this. First, certainly the embedding dimension $2b$ of the CLS tokens should be sufficiently large, otherwise we are facing the bottleneck already in the very beginning. Second, one can employ a third training phase. This motivates Phase 3: This third phase allows for a fine-tuning of all involved weights by jointly training the entire architecture starting from the pre-trained weights of the first two phases. To ensure stability of the learned representation space, this third (fine-tuning) step should be constrained by a strong regularization or it should only consider very few gradient descent steps with moderate step sizes (learning rates).

Clearly, to benefit from context data, the target batch and the context batch have to be associated. This association can reflect target-context pairs that consider the same products, the same local regions or the same circumstances in which these instances operate, without having these (soft) factors encoded in the features. Such examples are the most obvious ones in which the ICL-Credibility Transformer can successfully operate. A less obvious though similar set-up is the problem of having a high-cardinality categorical feature, e.g., vehicle models within a motor insurance portfolio. Categorical encoding (one-hot encoding, dummy coding or entity embedding) will then involve a vast number of parameters which cannot be estimated reliably (with credibility). Instead of trying to estimate a multitude of parameters, we rather could use similar instances as the context batch. Either there is a natural association or one can try to construct a neighborhood of similar instances using an unsupervised learning method; see \citet[Chapter 10]{AITools}.

Alternatively, one could also select a context batch that is not directly retrieval-related, but one rather uses this second training phase as a fine-tuning phase. That is, the base Credibility Transformer may not have learned the optimal CLS tokens because the early stopping may not only prevent from over-fitting, but it may also prevent from sufficient fine-tuning. The second training phase with the context-augmented information can then be seen as the fine-tuning phase that tries to find the structure that was not found by the base Credibility Transformer training with early stopping. This view can also be interpreted as a boosting step.

\section{B\"uhlmann credibility and interpretations}
\label{section interpretation}
We establish a theoretical connection between the ICL mechanism and linear \cite{buhlmann1967experience} credibility. In the ICL setting, we partition our data into context  and target batches,  ${\cal B} = {\cal B}_{\rm context} \cup {\cal B}_{\rm target}$. The ICL objective is to enhance predictions on the target batch by augmenting the available information by the context batch. There are multiple credibility mechanisms involved in this architecture. Besides the base Credibility Transformer, there is the obvious one giving the decorated tokens \eqref{decorated tokens}. The weights $v_i/(v_i + \kappa)$  in these decorated tokens precisely reflect linear \cite{buhlmann1967experience} credibility weights that assigns a higher credibility weight for bigger case weights $v_i$. E.g., in claims counts modeling this is used to balance the different exposure lengths between the instances for experience rating.  
The following proposition unveils a more hidden credibility mechanism.

\begin{prop}
\label{prop:icl_credibility}
The attention mechanism in the causal attention head 
$\bH=[\bh_1,\ldots, \bh_n]^\top \in \R^{n\times 2b}$, given in 
\eqref{causal self-attention head}, has the following credibility structure for every instance $i \in {\cal I}_{\rm target}$ of the target batch
\begin{equation}\label{credibility 19}
\bh_{i} =
a_{i,i} \, \boldsymbol{z}^{\rm FNN}_{V}\left(\widehat{\bc}^{\rm cred}(\bx_i)\right) + \sum_{j \in {\cal I}_{\rm context}} a_{i,j}\,
\boldsymbol{z}^{\rm FNN}_{V}\left(\widehat{\bc}^{\rm cred}(\bx_j)+ \frac{v_j}{v_j+\kappa} \bz^{\rm FNN1}(Y_j)\right),
\end{equation}
with attention weights $a_{i,j} \ge 0$ satisfying $a_{i,i}+\sum_{j \in {\cal I}_{\rm context}} a_{i,j}=1$, the specific structure is given in \eqref{Form credibility weights}, below.
\end{prop}

A consequence of this result is that the attention head performs a credibility weighting between instance $i$ specific information and the augmented context information. Note that this credibility step is the only step where targets and the context interact. In terms of explainability, we can extract the learned attention weights $(a_{i,j})_{j \in {\cal I}_{\rm context} \cup \{i\}}$ to indicate from which context instance we can learn the most for a specific instance $i$ of the target batch.

This explanation \eqref{credibility 19} looks rather convincing, however, this reasoning has a catch that makes it different from linear \cite{buhlmann1967experience} credibility. Namely, the responses
$(Y_{j})_{j \in {\cal I}_{\rm context}}$ also enter the attention weights
$(a_{i,j})_{j \in {\cal I}_{\rm context} \cup \{i\}}$; not being explicitly expressed in this notation. That is, in fact, we have a non-linear structure \eqref{credibility 19} in the responses $(Y_j)_{j \in {\cal I}_{\rm context}}$ and a linear \cite{buhlmann1967experience} credibility interpretation only reflects a first order approximation.
This point could be modified to a proper linear credibility formula by considering the changed query and key tensors (not depending on the responses)
\begin{eqnarray}\nonumber
\widetilde{\bQ} &=& \left[\boldsymbol{z}^{\rm FNN}_{Q}\left(\bc^{\red{\rm cred}}(\bx_i)\right)\right]_{i\in {\cal I}}^\top~\in~\R^{n \times 2b},\\\label{linear ICL Z}
\widetilde{\bK} &=& \left[\boldsymbol{z}^{\rm FNN}_{K}\left(\bc^{\red{\rm cred}}(\bx_i)\right)\right]_{i\in {\cal I}}^\top~\in~\R^{n \times 2b},\\\nonumber
\bV &=& \left[\boldsymbol{z}^{\rm FNN}_{V}\left(\bc^{\rm decor}(\bx_i)\right)\right]_{i\in {\cal I}}^\top~\in~\R^{n \times 2b},
\end{eqnarray}
i.e., changing the queries and keys to purely feature driven (excluding the responses), would give a proper linear credibility formula in the observations $Y_j$ in \eqref{credibility 19}. This option can easily be used if we have a single ICL Transformer layer, but cannot easily be adapted to deep ICL Transformers.
We will also consider this option \eqref{linear ICL Z} in our example below.

\bigskip

{\Beweis
{\bf Proof.} The proof is an easy consequence of the attention mechanism in the attention head \eqref{causal self-attention head}.
Focusing on the $i$-th row of the causal attention layer $\bA$, we have through the softmax operator
\begin{equation*}
a_{i,j} =  \frac{\exp\left( \bq_i^\top \bk_j/\sqrt{2b} + M_{i,j}^\infty \right)}
{\sum_{k=1}^n \exp\left( \bq_i^\top \bk_k/\sqrt{2b} + M_{i,k}^\infty \right)}~>~0,
\qquad \text{ for $j \in {\cal I}$,}
\end{equation*}
with query tensor $\bQ=[\bq_1,\ldots, \bq_n]^\top$ and key tensor
$\bK=[\bk_1,\ldots, \bk_n]^\top$.
The softmax operator normalizes this to 1, thus, $(a_{i,j})_{j=1}^n$ reflect normalized weights for all $i$. If we select an instance from the target set $i \in {\cal I}$, we receive through the masking operator $\bM$
\begin{equation}\label{Form credibility weights}
a_{i,j} =  \frac{\exp\left( \bq_i^\top \bk_j/\sqrt{2b} \right)}
{\exp\left( \bq_i^\top \bk_i/\sqrt{2b}\right) + \sum_{k \in {\cal I}_{\rm context}} \exp\left( \bq_i^\top \bk_k/\sqrt{2b} \right)}~\in ~(0,1),
\qquad \text{ for $j \in {\cal I}_{\rm context} \cup \{i\}$,}
\end{equation}
and $a_{i,j}=0$ if $j \in {\cal I}_{\rm target}\setminus \{i\}$. The latter ensures that there is no interaction within the target batch, and the remainder gives the classical credibility structure on ${\cal I}_{\rm context} \cup \{i\}$.
Inserting this into the attention head \eqref{causal self-attention head}
gives us for the target instances $i \in {\cal I}_{\rm target}$
\begin{eqnarray*}
\bh_{i} &=& \sum_{j \in {\cal I}} a_{i,j}  \bv_{j}~=~a_{i,i}  \bv_{i} + \sum_{j \in {\cal I}_{\rm context}} a_{i,j}  \bv_{j},
\\&=&
a_{i,i} \, \boldsymbol{z}^{\rm FNN}_{V}\left(\widehat{\bc}^{\rm cred}(\bx_i)\right) + \sum_{j \in {\cal I}_{\rm context}} a_{i,j}\,
\boldsymbol{z}^{\rm FNN}_{V}\left(\widehat{\bc}^{\rm cred}(\bx_j)+ \frac{v_j}{v_j+\kappa} \bz^{\rm FNN1}(Y_j)\right),
\end{eqnarray*}
for value tensor $\bV=[\bv_1,\ldots, \bv_n]^\top$. This completes the proof.
\EndProof}

\section{Numerical example}
\label{Numerical example}
We evaluate our approach on the well-known French MTPL dataset of \citet{Dutang}, which has been widely studied in the recent actuarial literature. The dataset contains policy-level information with claims frequencies, allowing us to assess the proposed ICL-Credibility Transformer model's ability to predict insurance claim rates. The characteristics of the dataset are summarized in Table \ref{tab:dataset}.\footnote{We use the data-cleaned version that can be downloaded from \url{https://aitools4actuaries.com/}.}

\begin{table}[h]
\footnotesize
\centering
\caption{Dataset characteristics}
\label{tab:dataset}
\begin{tabular}{lcc}
\toprule
\textbf{Characteristic} & \textbf{Training set} & \textbf{Test set} \\
\midrule
Number of policies & 610,206 & 67,801 \\
Total exposure (years) & 322,857 & 35,943 \\
Number of claims & 23,738 & 2,645 \\
Average frequency & 7.36\% & 7.35\% \\
\midrule
\multicolumn{3}{l}{\textbf{Covariate description}} \\
\midrule
\multicolumn{3}{l}{Categorical (4): Area, VehGas, VehBrand, Region} \\
\multicolumn{3}{l}{Continuous (5): VehPower, VehAge, DrivAge, BonusMalus, Density} \\
\multicolumn{3}{l}{Target: ClaimNb (claim count)} \\
\multicolumn{3}{l}{Exposure: Exposure (in yearly units)} \\
\bottomrule
\end{tabular}
\end{table}

The dataset exhibits typical characteristics of insurance portfolios: highly imbalanced with most policies having zero claims, heterogeneous exposure periods, and a mix of categorical and continuous covariates.

We use the Poisson deviance loss as our primary loss metric, consistent with actuarial practice and presenting a strictly consistent scoring function, see \citet{GneitingRaftery},
\begin{equation*}
L_{\text{Poisson}} = \frac{2}{n} \sum_{i=1}^{n} \left[\widehat{\mu}_i - Y_i - Y_i \log\left(\frac{\widehat{\mu}_i}{Y_i}\right)\right],
\end{equation*}
with $\widehat{\mu}_i=\widehat{\mu}(\bx_i)$ being the predictor for the claim $Y_i$ and where the square bracket is set equal to $\widehat{\mu}_i$ for $Y_i=0$.

We report on two main versions of the ICL-Credibility  Transformer. The first one is exactly the ICL-Credibility Transformer as described in the previous section with the non-linear credibility mechanism in the attention head, see \eqref{credibility 19}. The second one is the simpler linearized ICL-Credibility Transformer, where we restrict the queries and keys to be purely feature driven (excluding the responses), see \eqref{linear ICL Z}. For the former model, we utilize two ICL Transformer layers and for the latter model, we utilize one ICL Transformer layer, so as to preserve the linear credibility structure, as discussed above. The training will have three phases. 
(Phase 1) In the first phase, for both of the models, we perform a single training run on the base Credibility Transformer model. For this, we select the AdamW optimizer (learning rate $10^{-3}$, weight decay $10^{-2}$, $\beta_2=0.95$), batch size 1024, Poisson deviance loss, and a 15\% validation split with early stopping (patience 20). (Phase 2) Then, we freeze the Credibility Transformer's decoder $\widehat{\bz}^{\rm decod}$ and insert the outcome token decorator and the ICL Transformer layers between the CLS token and the output of the Credibility Transformer model. In this second training phase, we train the outcome token decorator, the ICL Transformer, the base Credibility Transformer's encoder and the initial embedding weights for 50 epochs with the AdamW optimizer (learning rate $3\cdot 10^{-4}$, weight decay $10^{-2}$, $\beta_2=0.95$) while maintaining causal masking so that target examples do not interact. Batches are constructed as a concatenation of a context set and a target set of examples, and we apply the loss function only on the target set via sample weights. (Phase 3) The final third phase performs a fine-tuning step of the entire architecture for 20 epochs, i.e., including all of the components of the ICL-Credibility Transformer (the base Credibility Transformer, the output token decorator, the ICL transformer and the decoder) with a small AdamW learning rate $3\cdot 10^{-5}$ and early stopping (patience 10).

The instances within the context batches are selected so as to be similar to the instances with the target batches. This is done by performing an approximate neighborhood search of the target instances within the context batch. For each step, we assemble batches as $\big[{\cal B}_{\rm context} \,\|\, {\cal B}_{\rm target}\big]$ with context size $c=\min(1000,\lvert{\cal D}_{\rm train}\rvert)$ and target chunk size $m=200$, i.e., we use a ratio of 5 context rows for each target row. At inference, we partition the test set into chunks of size $m=200$. For each test chunk, we compute base Credibility Transformer's encoder embeddings 
$\widehat{\bc}^{\rm cred}(\bx)$ for the $m$ test rows and retrieve their top-$(K=64)$ nearest neighbors from the training set (cosine via inner product on $\ell_2$-normalized vectors). That is, the context batch ${\cal B}_{\rm context}$
is retrieved from the base Credibility Transformer model's CLS token embedding space; we interpret similarity in the credibilitized CLS token $\widehat{\bc}^{\rm cred}(\bx)$ as similarity in risk behavior.
By default, we use cosine similarity implemented as inner product on $\ell_2$-normalized embeddings. We query $K=64$ nearest neighbors per target example and then build a unique candidate pool by taking the union across the target chunk and ranking candidates by similarity. After ranking by similarity, we take the top $c = 1000$ candidates\footnote{In more detail, for each test chunk of size $m$, we retrieve the top-$K$ nearest training neighbors per test row in the base embedding space (cosine via inner product on $\ell_2$-normalized vectors). Let $\mathcal{N}(x_r)$ denote the $K$ neighbors of test point $x_r$ and define the candidate pool $\mathcal{C}=\bigcup_{r=1}^m \mathcal{N}(x_r)$. We assign each candidate $j\in\mathcal{C}$ a single score given by its best match over the chunk, namely $s_j=\max_r \mathrm{sim}(x_r,j)$ for cosine similarity. We then sort the candidates by $s_j$ in descending order for the cosine similarity, break ties deterministically, and select the first $c=1000$ unique items as the context set.}. For this, we employ the FAISS package \citep{johnson2019billion} for high-throughput nearest neighbor search.  
This forms the batches $\big[{\cal B}_{\rm context}\,\|\,{\cal B}_{\rm target}\big]$, we supply the observed responses only for ${\cal B}_{\rm context}$ (zeros for ${\cal B}_{\rm target}$), and apply a causal mask that prevents target-target interactions while allowing target-context attention. A single forward pass yields predictions for the target block; we keep the last $m$ outputs and iterate over all chunks to cover the full test set. 
For a summary, see Table \ref{tab:impl_hparams}.

\begin{table}[h]
\footnotesize
\centering
\caption{Key implementation hyper-parameters (ICL)}
\label{tab:impl_hparams}
\begin{tabular}{ll}
\toprule
\textbf{Component} & \textbf{Setting} \\
\midrule
Base CT pre-training (phase 1) & AdamW (LR $10^{-3}$, WD $10^{-2}$, $\beta_2=0.95$), 100 epochs, batch 1024 \\
ICL fine-tuning (phase 2) & AdamW (LR $3\!\cdot\!10^{-4}$, WD $10^{-2}$, $\beta_2=0.95$), 50 epochs \\
Joint fine-tuning (phase 3) & AdamW (LR $3\!\cdot\!10^{-5}$), 20 epochs \\
Objective & Poisson deviance loss \\
Validation & Early stopping: base CT pre-train 20, ICL 20, joint 10 (patience) \\
Batching & Context $c=1000$, target chunk $m=200$; loss on target rows only \\
Retrieval metric & Cosine similarity \\
Neighbors & $K=64$ per target; union, optional per-target quota, cap at $c$ \\
Pre-computation & Train$\to$train neighbors (default on); optional val/test$\to$train; on-disk cache \\
\bottomrule
\end{tabular}
\end{table}

We compare the ICL-Credibility Transformer against several baselines taken from \citet{WM2023}, \citet{Brauer}, \citet{richman2024credibility} and \citet{richman2025tree}. Table \ref{main_results} presents the out-of-sample performance comparison across all models.

\begin{table}[htb!]
  \centering
  {\footnotesize
  \begin{center}
  \begin{tabular}{|l||c|cccc|}
  \hline
  & \# &\multicolumn{2}{c}{In-sample\,\,}&\multicolumn{2}{c|}{Out-of-sample\,\,}\\
  Model & Param.&\multicolumn{2}{c}{Poisson loss\,\,} & \multicolumn{2}{c|}{Poisson 
    loss\,\,} \\
  \hline\hline
  Null model (intercept-only) &1&25.213& &25.445&\\
  Poisson GLM3 &50& 24.084 && 24.102&\\
  Poisson GAM &  (66.7) &23.920&&23.956&\\\hline
  Plain-vanilla FNN &792& 23.728 & ($\pm$ 0.026)& 23.819&($\pm$ 0.017)\\
  Ensemble plain-vanilla FNN &792& 23.691 && 23.783&\\\hline
  CAFTT & 27,133& 23.715 &($\pm$ 0.047) & 23.807 &($\pm$ 0.017)\\
  Ensemble CAFTT & 27,133& 23.630 & & 23.726 &\\\hline
  Credibility Transformer & 1,746& 23.641 &($\pm$ 0.053) & 23.788 &($\pm$ 0.040)\\
  Ensemble Credibility Transformer & 1,746& 23.562 && 23.711 &
  \\ \hline
  Tree-like PIN & 4,147& 23.593 &($\pm$ 0.046) & {23.740} &($\pm$ 0.025)\\ 
  Ensemble Tree-like PIN & 4,147& 23.522 && {23.667} &\\ \hline\hline
  Single run base Credibility Transformer (phase 1)& 15,614 & 23.653 & & 23.743 &\\\hline
  ICL Transformer - 2 Layers (phase 2) & 46,439 & 23.631 & ($\pm$ 0.048) & 23.725 & ($\pm$ 0.022) \\
  Ensemble ICL Transformer - 2 Layers (phase 2) & 46,439& 23.584 && 23.679
 &\\
  ICL Transformer Fine-tuning (phase 3) & 46,439& 23.561 & ($\pm$ 0.024) & 23.710 & ($\pm$ 0.009) \\
Ensemble ICL Transformer Fine-tuning (phase 3) & 46,439& 23.521 && 23.676
&\\\hline
  ICL Transformer Linearized (phase 2)& 33,199& 23.634 & ($\pm$ 0.046) & 23.766 & ($\pm$ 0.034) \\
  Ensemble ICL Transformer Linearized (phase 2)& 33,199& 23.570 && 23.699 &\\
  ICL Transformer Linearized Fine-tuning (phase 3)& 33,199& 23.563 &($\pm$0.043) & 23.723 & ($\pm$ 0.041) \\
  Ensemble ICL Transformer Linearized Fine-tuning (phase 3)& 33,199& 23.514
 && 23.678
 &\\\hline
  \end{tabular}
  \end{center}}
  \caption{Number of parameters, in-sample and out-of-sample Poisson deviance losses (units are in $10^{-2}$). Benchmark models for Null model through Poisson GAM are taken from \citet[Table 7.9]{WM2023}; plain-vanilla FNN and ensemble results are from \citet[Table 7.9]{WM2023}; CAFTT results are from \citet[Tables 2 and 4]{Brauer}; Credibility Transformer results are from \citet[Table 2]{richman2024credibility}; Tree-like PIN results are from \citet[Table 2]{richman2025tree}.}
  \label{main_results}
\end{table}

Generally, because early stopped stochastic gradient descent fitting involves several items of randomness, e.g., through the random initializations of the algorithms, we perform 5 fitting runs for all architectures, and the shown results are averages over these 5 fitting runs; in brackets we provide the observed standard deviations.
The single run of the base Credibility Transformer produces a result of $23.743$ (out-of-sample), which is roughly within 1 standard deviation of the results obtained in the original Credibility Transformer model $23.788 \pm 0.040$. Adding the context token decorator and the ICL Transformer to the base Credibility Transformer model produces small out-of-sample improvements for the version with two layers $23.725$; however, the linearized version yields $23.766$ (out-of-sample), demonstrating a slight worsening in performance. Fine-tuning all parameters of the ICL-Credibility Transformer produces further small improvements, with the fine-tuned non-linear ICL-Credibility Transformer version achieving the best test performance $23.710$ compared to $23.723$ for the linearized version. This latter point means that the linearized version benefits from ICL once full fine-tuning is performed. The scores discussed thus far for the linearized and two-layer models are the average scores of 5 independent fitting runs (as mentioned at the beginning of this paragraph).

The best scores are achieved after ensembling (averaging) the predictions of the 5 independent fitting runs together, that is, we average the predictions of the 5 fitting runs and we evaluate the performance of this average (ensembled) prediction. The scores improve after ensembling the 5 models, with the linearized model achieving a score of $23.699$ before fine-tuning and $23.678$ after fine-tuning. The best out-of-sample ICL results are achieved by the two-layer model, with $23.679$ pre-fine-tuning and $23.676$ post-fine-tuning. Notably, the simpler linearized model performs almost as well as the larger non-linear model after fine-tuning.

We observe that the ICL versions have quite significantly improved upon the results of the original (base) Credibility Transformer. This affirmatively answers the opening questions at the beginning of this paper, whether ICL can improve predictive models, and this means that the ICL Transformer approach taken in this paper can improve on supervised learning for actuarial applications! 

\medskip

We now analyze how applying the steps taken to build the ICL-Credibility Transformer affects the CLS tokens and the predictions. For simplicity, we select a single run of the ICL-Credibility Transformer training, and we perform a principal component analysis (PCA) of the CLS tokens taken from the base Credibility Transformer (i.e., before we have re-trained). The PCA is applied to the following tokens:
\begin{itemize}
\item[(1)] CLS tokens of the base Credibility Transformer  before modifying this model (training phase 1);
\item[(2)]
  (2a)  CLS tokens,  (2b) decorated CLS tokens before the ICL mechanism, (2c) decorated CLS tokens after performing ICL augmentation (these are shown after training phase 2);
\item[(3)] (3a)  CLS tokens,  (3b) decorated CLS tokens before the ICL mechanism, (3c) decorated CLS tokens after performing ICL augmentation (these are shown again after training pahase 3, i.e., fine-tuning the ICL-Credibility Transformer end-to-end).
\end{itemize}
Because we use the identical PCA for all three models (phases 1-3), we can directly compare the projections of the CLS tokens as these are being modified. 

\begin{figure}[h]
\centering
\includegraphics[width=.75\textwidth]{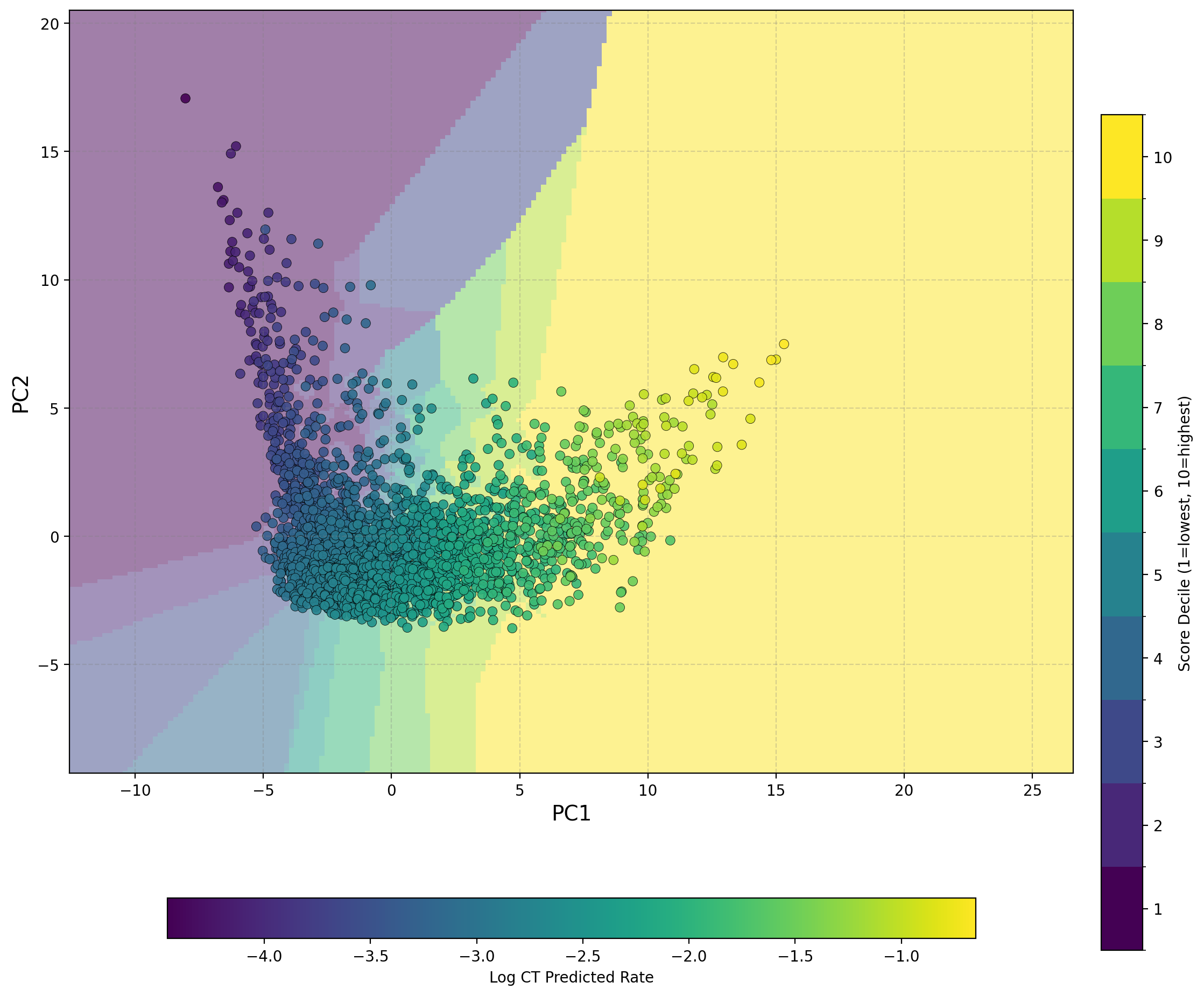}
\caption{2D PCA representation of the CLS tokens from the base Credibility Transformer (phase 1), before being used in the ICL model architecture. The background of the figure represents a color shade depending on the points in that region using a Voronoi-like tessellation.}
\label{pca_icl_ct}
\end{figure}

Phase 1: Figure \ref{pca_icl_ct} shows the results of the base Credibility Transformer's predictions for 3,000 randomly selected policies in the test set. To better understand the behavior we color code the points based on the predicted log-frequency (rate) of the policies. We then apply a Voronoi-like tessellation\footnote{ A Voronoi-like tessellation divides the plane into numerous smaller regions, each filled with a color representing its closest data point.} to visualize how the base Credibility Transformer adjusts the risk zones.
From Figure~\ref{pca_icl_ct}, we can see that the base Credibility Transformer places its perceived low-risk policies in the top-left quadrants and its perceived high-risk policies in the right-half of the plane.

\begin{figure}[h]
\centering
\includegraphics[width=1\textwidth]{first_fine_tune.png}
\caption{2D PCA representation of the (2a) CLS tokens from the base Credibility Transformer after phase 2, (2b) decorated CLS tokens before the ICL mechanism and (2c) decorated CLS tokens after performing ICL augmentation.}
\label{phase_1}
\end{figure}


Phase 2: After transferring the base Credibility Transformer model to the ICL model and retraining it with the decoder frozen, we obtain the PCA plots shown in Figure~\ref{phase_1}. The CLS token representations, which previously appeared as in Figure~\ref{pca_icl_ct}, are now condensed or almost squashed together, as seen in Figure~\ref{phase_1}(a). This occurs because the base model no longer needs to encode all discriminative information itself; it can be viewed as the base transformer learning to function as part of a pipeline rather than standalone.
A pattern then begins to emerge in Figure~\ref{phase_1}(b), after the decoration stage, which is largely driven by a linear projection of the embeddings. With more degrees of freedom, these two plots taken together can be interpreted as the base deferring structural differentiation to the projection layer. Importantly, the shape produced in Figure~\ref{phase_1}(b) differs from that in Figure~\ref{pca_icl_ct}, implying that it is not simply undoing the "condensation" from (a) but rather preparing the representations for the ICL transformer.

Finally, in Figure~\ref{phase_1}(c), the points show slightly more refined adjustments. Test points attend to labeled context rows (training examples with outcome decorations), so attention pulls outcome information from similar training examples, nudging test point positions based on evidence from labeled neighbours. These small nudges make sense because the projection layer in (b) already creates a good structure; the ICL attention then makes targeted adjustments rather than complete restructuring, simply aligning cases with what has happened to similar cases. The quadrants of the Voronoi-like tessellation from Figure \ref{phase_1} display a more defined split between the low- and high-risk policies than those relative to the base Credibility Transformer from Figure \ref{pca_icl_ct}; recall that the decoder $\widehat{\bz}^{\rm decod}$ is the same in both figures.

\begin{figure}[h]
\centering
\includegraphics[width=1\textwidth]{second_fine_tune.png}
\caption{2D PCA representation of the (3a) CLS tokens from the base Credibility Transformer after phase 3, (3b) decorated CLS tokens before the ICL mechanism and (3c) decorated CLS tokens after performing ICL augmentation.}
\label{phase_2}
\end{figure}

Phase 3:
After fine-tuning the entire model - unfreezing the decoder that was frozen in phase 2 - we obtain PCA plots very similar to those produced after phase 2. We observe similar patterns between Figure~\ref{phase_1} and Figure~\ref{phase_2}, implying that the performance uplift comes primarily from adjusting the decoder to better exploit the more refined representations provided by the ICL transformer. By adapting to this new information, the decoder improves the model's overall predictive power.

Figures~\ref{first_phase_differences} and~\ref{second_phase_differences} demonstrate that the ICL process does not make significant adjustments to all points. Rather, it seems to select the points that differ from the rest of their context and adjust those points based on the points surrounding context.

\begin{figure}[h]
\centering
\includegraphics[width=1\textwidth]{first_fine_tune_differences.png}
\caption{2D PCA representation of the (4a) CLS tokens from the base Credibility Transformer after phase 2, (4b) decorated CLS tokens before the ICL mechanism and (4c) decorated CLS tokens after performing ICL augmentation, where the colors of the points represent the adjustment made on the log-scale to the base Credibility Transformer rate predictions by the ICL process.}
\label{first_phase_differences}
\end{figure}

\begin{figure}[h]
\centering
\includegraphics[width=1\textwidth]{second_fine_tune_differences.png}
\caption{2D PCA representation of the (5a) CLS tokens from the base Credibility Transformer after phase 3, (5b) decorated CLS tokens before the ICL mechanism and (5c) decorated CLS tokens after performing ICL augmentation, where the colors of the points represent the adjustment made on the logscale to the base Credibility Transformer rate predictions by the ICL process.}
\label{second_phase_differences}
\end{figure}

\section{Zero-shot analysis using ICL}
\label{Zero-shot analysis of Region}
\subsection{Motivation for zero-shot prediction}
{\it Zero-shot prediction} is used here to refer to the ability of a  model to make accurate inferences on categories or feature levels that were entirely absent during model training. In practical insurance applications, this capability can be crucial, as new categories may emerge over time -- for example, a new vehicle model introduced by a manufacturer or an insurance product expanded to a previously uncovered geographical region. Traditional models often struggle with such unseen categories, thus, in practice, often one defaults to simple imputation of a mean or median value, or setting the categorical level to a ``missing" indicator. In contrast, the ICL mechanism of the ICL-Credibility Transformer architecture enables the model to leverage contextual similarities from related instances, potentially generalizing to these novel cases without retraining the model; it is exactly this ability that we test in this section.

\subsection{Set-up for zero-shot prediction}

In the following example, we focus on the `Region' categorical feature from the French MTPL dataset, and we design a zero-shot experiment as follows. We create a new train-test split where certain regions are treated as completely ``unseen'' (unavailable during model training). Specifically, we select records comprising approximately 10\% of the exposure of the entire dataset, corresponding to the regions with the lowest amounts of exposures, these are ${\cal R}_{\rm new}=\{$'R43', 'R21', 'R42', 'R94', 'R83', 'R74', 'R23', 'R22', 'R26', 'R25', 'R73'$\}$. These regions are designated as the new test set, and their labels are remapped to a new ``unseen'' category to simulate novelty during inference. For the remaining training data, we further set additional region levels to ``unseen'' to teach the original base Credibility Transformer how to handle this category, i.e., to provide a parameter estimate for the embedding level corresponding to ``unseen''. For this, we select the next smallest regions in terms of exposures $\{$'R41', 'R54', 'R31', 'R72', 'R91', 'R52'$\}$, which account for roughly 20\% of the entire exposure of the training sample. Thus, we ensure that the model encounters the ``unseen'' label during model training, learning to rely on contextual information from similar instances in these cases. The full ICL-Credibility Transformer model pipeline is then trained on this modified dataset, and its predictive performance is assessed on the (hold-out) test set, quantifying its zero-shot generalization to truly novel regions ${\cal R}_{\rm new}$. Table \ref{tab:dataset_split} summarizes the key information of the train-test split. Interestingly, we see that the frequency in the new test set is a lower than in the training set; this is due to the new manner in which we have performed the split above.

\begin{table}[h]
\footnotesize
\centering
\label{tab:dataset_split}
\begin{tabular}{lcc}
\toprule
\textbf{Characteristic} & \textbf{Training set} & \textbf{Test set} ${\cal R}_{\rm new}$\\
\midrule
Number of policies & 601,781 & 76,226 \\
Number of records set to "unseen" & 165,200 & 76,226 \\
Total exposure (years) & 323,458 & 34,900 \\
Number of claims & 24,006 & 2,377 \\
Average frequency & 7.42\% & 6.81\% \\
\midrule
\multicolumn{3}{l}{\textbf{Feature description}} \\
\midrule
\multicolumn{3}{l}{Categorical (4): Area, VehGas, VehBrand, Region/``unseen''} \\
\multicolumn{3}{l}{Continuous (5): VehPower, VehAge, DrivAge, BonusMalus, Density} \\
\multicolumn{3}{l}{Target: ClaimNb (claim count)} \\
\multicolumn{3}{l}{Exposure: Exposure (in years)} \\
\bottomrule
\end{tabular}
\caption{Characteristics of the training and test sets for the zero-shot analysis.}
\end{table}

\subsection{Regional analysis}

As a preliminary introduction to better understand the performance of our zero-shot approach across different regions, we present a detailed breakdown of the results by region in Table \ref{tab:regional_analysis}. This table shows the claim counts, exposures, claim rates, and Poisson deviances for each region, along with whether it was designated as ``unseen'' and its assignment to the training or test set. The Poisson deviances have been calculated by using the average observed frequencies within each region as the predictions.

\begin{table}[htb!]
\centering
{\footnotesize
\begin{tabular}{lrrrccl}
\toprule
\textbf{Region} & \textbf{ClaimNb} & \textbf{Exposure} & \textbf{Claim Rate} & \textbf{Poisson Dev.} & \textbf{Unseen} & \textbf{Set} \\
\midrule
R43 & 38 & 564 & 0.07 & 0.215 & yes & test \\
R21 & 77 & 1,204 & 0.06 & 0.179 & yes & test \\
R42 & 92 & 1,209 & 0.08 & 0.273 & yes & test \\
R94 & 132 & 1,766 & 0.07 & 0.210 & yes & test \\
R83 & 141 & 2,322 & 0.06 & 0.187 & yes & test \\
R74 & 197 & 2,396 & 0.08 & 0.268 & yes & test \\
R23 & 220 & 3,177 & 0.07 & 0.178 & yes & test \\
R22 & 314 & 3,573 & 0.09 & 0.246 & yes & test \\
R26 & 345 & 5,023 & 0.07 & 0.219 & yes & test \\
R25 & 452 & 6,653 & 0.07 & 0.263 & yes & test \\
R73 & 369 & 7,014 & 0.05 & 0.158 & yes & test \\
\midrule
R41 & 468 & 8,112 & 0.06 & 0.241 & yes & train \\
R54 & 800 & 11,163 & 0.07 & 0.271 & yes & train \\
R31 & 944 & 11,488 & 0.08 & 0.227 & yes & train \\
R72 & 1,055 & 14,316 & 0.07 & 0.222 & yes & train \\
R91 & 1,007 & 14,709 & 0.07 & 0.198 & yes & train \\
R52 & 1,576 & 21,930 & 0.07 & 0.263 & yes & train \\
\midrule
R53 & 1,871 & 27,753 & 0.07 & 0.282 & no & train \\
R11 & 2,591 & 30,198 & 0.09 & 0.238 & no & train \\
R93 & 2,986 & 35,749 & 0.08 & 0.240 & no & train \\
R82 & 4,233 & 45,333 & 0.09 & 0.300 & no & train \\
R24 & 6,475 & 102,706 & 0.06 & 0.266 & no & train \\
\bottomrule
\end{tabular}
}
\caption{Regional breakdown: Poisson deviances by Region are computed based on the sample frequency in the corresponding Region.}
\label{tab:regional_analysis}
\end{table}

The regional analysis reveals interesting patterns in the distribution of risk across the different geographical areas. The test regions (all designated as ``unseen'') show varying frequency rates ranging from 0.05 to 0.09, with corresponding Poisson deviances between 0.158 and 0.273. Notably, regions with higher claim frequencies tend to have higher Poisson deviances. The training regions show a similar pattern, with the ``unseen'' training regions having lower exposures than the ``seen" regions (by construction). These ``seen" regions have a notably higher set of claims frequencies and Poisson deviances than the ``unseen'' test set.

\begin{table}[htb!]
\centering
{\footnotesize
\begin{tabular}{lc}
\toprule
\textbf{Description} & \textbf{Poisson Deviances} \\
\midrule
Whole portfolio & 0.255 \\
Test (``unseen'') & 0.216 \\
Train - ``unseen'' & 0.238 \\
Train - region provided & 0.267 \\
\bottomrule
\end{tabular}
}
\caption{Weighted average Poisson deviances across the different portfolio segments.}
\label{tab:weighted_poisson}
\end{table}

To provide a comprehensive view of the model performance across the different data segments, Table \ref{tab:weighted_poisson} presents the exposure-weighted average Poisson deviances for the key portfolio segments.
The weighted averages show that the test set achieves the lowest average Poisson deviance (0.216), followed by the ``unseen'' training regions (0.238). Interestingly, the ``seen" training regions with full regional information show the highest average Poisson deviance (0.267).
The significant differences between train and test, and ``seen" and ``unseen" should be kept in mind when understanding the zero-shot results presented in the next section, in particular, the differences between in-sample and out-of-sample results.

\subsection{Zero-shot results}
To evaluate the performance of the ICL-Credibility Transformer model on the ``unseen'' data (regions), the model is first trained on the training sample. When reporting the results of all of the modeling steps, we filter the training set to assess the performance only on records labeled ``unseen'' regions. Likewise, the entire test set consists of exactly these ``unseen'' records, but for a different set of regions ${\cal R}_{\rm new}$, as described above. We follow the training approach outlined in Section~\ref{Numerical example}: (1) first, we fit a base Credibility Transformer; (2) then, we add 2 ICL layers and freeze the decoder while training for 50 epochs; (3) finally, in the third phase, we unfreeze all weights and train for an additional 20 epochs. All other training hyper-parameters are identical to those in Table~\ref{tab:impl_hparams}. The results are displayed in Table \ref{main_results2}.

\begin{table}[htb!]
  \centering
  {\footnotesize
  \begin{center}
  \begin{tabular}{|l||c|cc|}
  \hline
  & \# &\multicolumn{1}{c}{In-sample ``unseen''}&\multicolumn{1}{c|}{Out-of-sample}\\
  Model & Param.&\multicolumn{1}{c}{Poisson loss} & \multicolumn{1}{c|}{Poisson
    loss on ${\cal R}_{\rm new}$}\\
  \hline\hline
  Null model (intercept-only) &1&23.464 &21.091\\\hline
  base Credibility Transformer (phase 1)& 15,294 & 22.275  & 20.282 \\\hline
  ICL-Credibility Transformer  (phase 2) & 46,119 & 22.315 & 20.264  \\
  Fine-tuning ICL-Credibility Transformer  (phase 3) & 46,119& 22.298 & 20.259
  \\\hline
  \end{tabular}
  \end{center}}
  \caption{Number of parameters, in-sample and out-of-sample Poisson deviance losses (units are in $10^{-2}$). Null model has been calculated to be used as a baseline using the observed frequency of the training data.}
  \label{main_results2}
\end{table}

Given that this training-test split deviates from the more conventional set-up shown before, we fit a null model (intercept-only) as a baseline. The base Credibility Transformer significantly outperforms the null model on both in-sample and out-of-sample Poisson deviance losses, as we would expect. Adding ICL mechanism to the base Credibility Transformer (with the frozen decoder) further improves the out-of-sample loss from $20.282$ to $20.264$. This gain highlights the benefits of cross-record attention enriched with contextual information, as well as the role of the credibility weighting, which allows tokens to enhance their predictive power by attending to other contextually similar tokens in the context batch. Finally, fine-tuning the full ICL-Credibility Transformer end-to-end yields an additional out-of-sample improvement to $20.259$. This indicates that the model benefits from adapting its decoder to the newly enriched context provided by ICL to the CLS tokens.

Thus, we observe that the ICL-Credibility Transformer generalizes fairly well to unseen the levels, provided that it receives some training exposure to handling such unseen levels. In our example, we sacrifice information on a few low-exposure levels to allow the model to learn how to treat these missing levels. However, this approach does result in a loss of data. Alternative methods for including these levels might involve creating duplicates of records designated as ``unseen'' while retaining both the original and the ``unseen'' labels. Care must be taken, however, to ensure that these training records do not serve as context for one another, as this could cause the model to perform worse on truly unseen test cases, i.e., such a leakage of information must be avoided for successful model training.

\section{Discussion and conclusions}
\label{Conclusions}
In this paper, we introduced the ICL enhanced Credibility Transformer, an architecture, which is, to our knowledge, a novel approach within the actuarial literature. This new model integrates the dynamic, example-driven learning paradigm of ICL with the robust, regularized representation learning of the Credibility Transformer. We have demonstrated empirically on the French MTPL dataset that this approach yields significant improvements in predictive accuracy, outperforming the relatively strong baseline Credibility Transformer upon which we build this new model.

A key insight of our work is the formal connection we established between the ICL mechanism and classical credibility theory. As shown in Proposition \ref{prop:icl_credibility}, the cross-attention layer at the heart of our ICL module functions as a sophisticated, data-driven credibility weighting scheme. The attention weights, which determine the influence of each context instance, can be interpreted as adaptive credibility factors that generalize the linear, variance-based weights of B\"uhlmann credibility to a non-linear, high-dimensional setting learned directly from feature interactions. Our qualitative analysis, supported by PCA and nearest-neighbor examinations, confirmed this interpretation. We observed that the ICL process refines the CLS token representations, pulling policies into more actuarially coherent clusters based on shared risk characteristics, thereby enabling more nuanced and accurate predictions.

Furthermore, the zero-shot learning experiment highlighted a particularly powerful practical advantage of the ICL-Credibility Transformer: its ability to generalize to new, unseen feature levels. By providing context from known risks that share other relevant features, the model can make better predictions for novel categories -- such as a new vehicle brand or a geographical region not present in the training data -- without any retraining. This capability addresses a common and persistent challenge in real-world insurance pricing, where portfolios are dynamic and new risks constantly emerge.

Looking forward beyond this initial exploration of ICL within an actuarial context, several practical issues remain. The performance of our model depends on the quality and relevance of the context batch. From a process perspective, while we employed a nearest-neighbor search in the CLS token space, scaling this retrieval process to real world portfolios may present a computational challenge. More importantly, a critical hurdle for a practical adoption is the justification of this dynamic, context-driven approach. Regulators will demand transparency and fairness, asking why a specific set of policies was used as context for a given individual. This requires developing a way to ensure this context does not introduce biases and results in premium rates that are actuarially sound. For policyholders, the rationale must be translatable into an explanation that their rate is fine-tuned using recent, relevant, and anonymized claims experience from a peer group to ensure maximum accuracy and fairness.

Another potential actuarial field of applications of ICL is individual claims reserving. In that case, the context can be formed by similar individual claims that are already further developed to gain more insight how a specific individual claim may evolve. In this context, there are less regulatory concerns about fairness and explainability because this mainly affects internal  actuarial processes that do not directly impact individual pricing of insurance contracts.

\bigskip

{\bf Declaration.} We have used ChatGPT to support us in the literature overview and to improve language and readability.

\bibliographystyle{apalike}
\bibliography{ref.bib}

\end{document}